\documentclass[a4paper]{article}
\usepackage[T1]{fontenc}
\usepackage{INTERSPEECH2019}
\usepackage{mathptmx}
\usepackage{stmaryrd}
\usepackage{mathtools}
\usepackage{amsmath}
\usepackage{graphicx}
\usepackage[noend]{algpseudocode}
\usepackage{url}
\usepackage{booktabs}
\usepackage[table]{xcolor}
\definecolor{light-gray}{gray}{0.9}
\usepackage{microtype}

\title{Sampling from Stochastic Finite Automata with Applications to
  CTC Decoding}
\name{Martin Jansche, Alexander Gutkin}
\address{Google Research, London, United Kingdom}
\email{\{mjansche,agutkin\}@google.com}

\begin{document}

\maketitle
\begin{abstract}
  Stochastic finite automata arise naturally in many language and speech
  processing tasks. They include stochastic acceptors, which represent
  certain probability distributions over random strings. We consider the
  problem of efficient sampling: drawing random string variates from the
  probability distribution represented by stochastic automata and
  transformations of those. We show that path-sampling is effective and can
  be efficient if the epsilon-graph of a finite automaton is acyclic. We
  provide an algorithm that ensures this by conflating epsilon-cycles
  within strongly connected components. Sampling is also effective in the
  presence of non-injective transformations of strings. We illustrate this
  in the context of decoding for Connectionist Temporal Classification
  (CTC), where the predictive probabilities yield auxiliary sequences which
  are transformed into shorter labeling strings. We can sample efficiently
  from the transformed labeling distribution and use this in two different
  strategies for finding the most probable CTC labeling.
\end{abstract}
\noindent\textbf{Index Terms}:
finite-state techniques,
stochastic automata,
sampling,
decoding

\section{Foundations}
\label{sec:intro}

Stochastic finite automata are used in many language and speech processing
tasks, including speech recognition and synthesis, language and
pronunciation modeling, tagging and parsing, among many others. They arise
naturally in optimization and decoding tasks for probabilistic sequence
models.

\subsection{FST background}
\label{sec:background}

A weighted finite state transducer (WFST) is a tuple $(\Gamma, \Delta, S,
Q, q_0, f, E, w)$ where $\Gamma$ and $\Delta$ are finite sets (the input
and output alphabet, respectively), $(S,\oplus,\otimes)$ is a semiring of
weights, $Q$ is a finite set of states, $q_0\in Q$ is the initial state,
$f: Q\to S$ is a function mapping a state to its final weight, $E\subseteq
Q\times(\Gamma\cup\{\varepsilon\})\times(\Delta\cup\{\varepsilon\})\times
Q$ is a finite set of edges, and $w: E\to S$ is a function mapping an edge
to its weight. Given an edge $e=(s,i,o,t)$ define canonical projections
$\mathrm{src}(e)=s$, $\mathrm{istr}(e)=i$, $\mathrm{ostr}(e)=o$, and
$\mathrm{tgt}(e)=t$.

A path $\pi=(e_1, \ldots, e_n)$ is a finite sequence of consecutive edges
such that $\mathrm{tgt}(e_i)=\mathrm{src}(e_{i+1})$ for
$i\in\{1,\ldots,n-1\}$.  In this case we write
$\mathrm{src}(\pi)=\mathrm{src}(e_1)$ and
$\mathrm{tgt}(\pi)=\mathrm{tgt}(e_n)$, and we extend the weight function to
paths:
\vspace*{-2mm}
\begin{displaymath}
  w(\pi) = \left[ \bigotimes_{i=1}^n w(e_i) \right] \otimes f(\mathrm{tgt}(\pi)) \, .
  \vspace*{-2mm}
\end{displaymath}
Associated with a path $\pi=(e_1,\ldots,e_n)$ is an input string
$\mathrm{istr}(\pi)=\mathrm{istr}(e_1) \cdots \mathrm{istr}(e_n)$ as well
as a similarly defined output string. The set of all successful paths
labeled with given strings $u\in\Gamma^*$ and $v\in\Delta^*$ is
$\mathrm{Paths}(u, v) = \{\pi \mid \mathrm{istr}(\pi)=u \land
\mathrm{ostr}(\pi)=v \land \mathrm{src}(\pi)=q_0\}$. The behavior of the
WFST $\mathcal{A}$ is the function $\llbracket\mathcal{A}\rrbracket:
\Gamma^* \times \Delta^* \to S$, and its grand total weight is
$w(\mathcal{A})$:
\begin{equation}\label{eq:prob}
  \llbracket\mathcal{A}\rrbracket(u, v) =
  \bigoplus_{\mathclap{\pi\in\mathrm{Paths}(u,v)}} w(\pi)
  \hspace*{2em}
  w(\mathcal{A}) =
  \bigoplus_{\mathclap{(u,v)\in\Gamma^*\times\Delta^*}}
  \llbracket\mathcal{A}\rrbracket(u,v) \, .
\end{equation}

\begin{figure}
  \centering%
  \includegraphics[clip,trim={36 36 36 36},height=2cm]{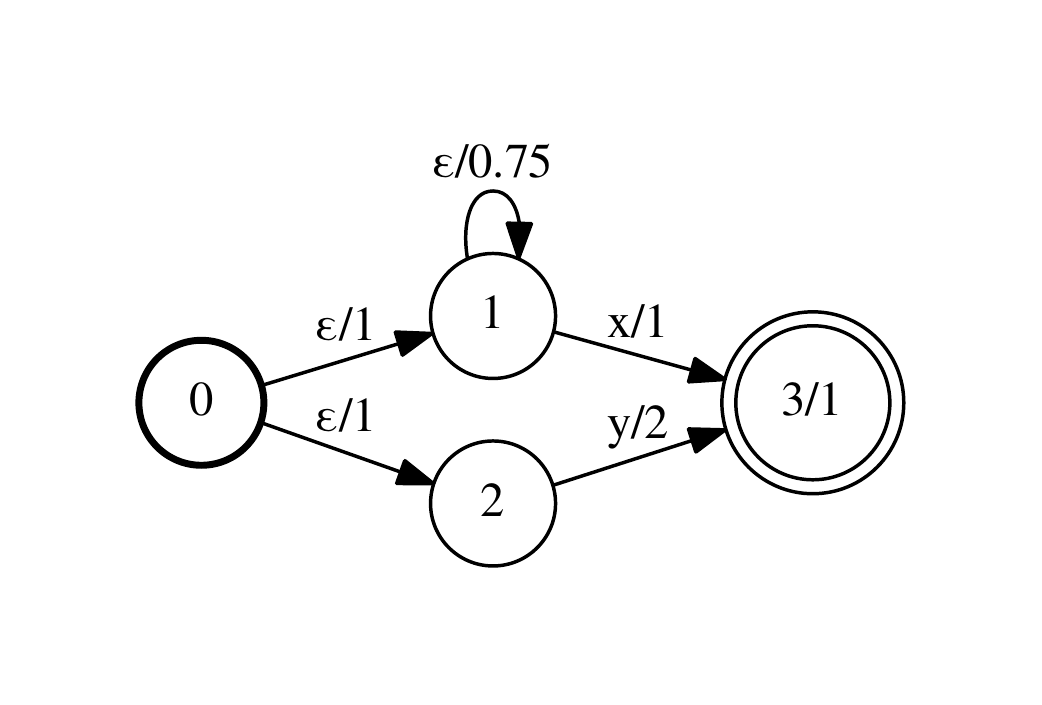}%
  \hspace*{\fill}%
  \includegraphics[clip,trim={36 36 36 36},height=2cm]{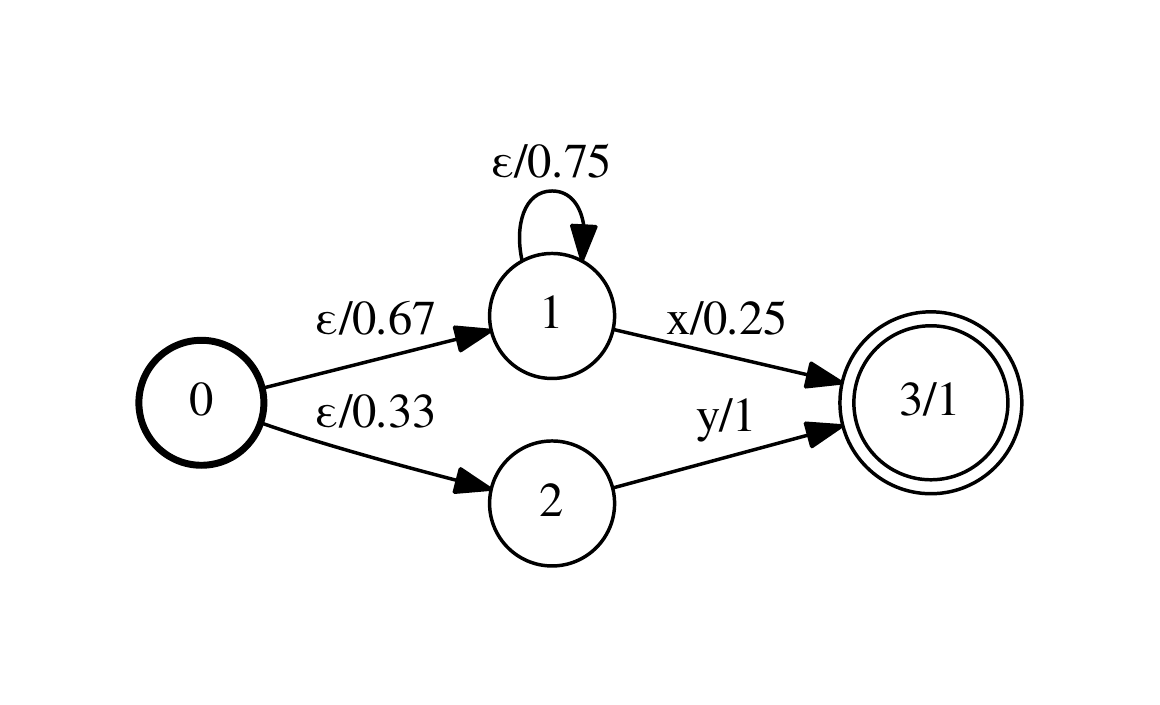}%
  \vspace{-2ex}%
  \caption{Unnormalized and equivalent normalized SFST.\\
    On the left, the grand total weight is $6$; on the right, $1$.}%
  \label{fig:push}%
  \vspace{-2mm}%
\end{figure}

A stochastic FST (SFST) is a WFST defined over the semiring of nonnegative
real numbers $(\mathbb{R}^{\geq0}, +, \cdot)$. An SFST $\mathcal{A}$ is
globally normalized if $w(\mathcal{A})=1$; in that case
$\llbracket\mathcal{A}\rrbracket$ is a probability measure for the
countable sample space $\Gamma^*\times\Delta^*$. Normalization is
impossible if the grand total diverges, due to cycles with path weights
$\geq1$. An SFST is locally normalized if
\begin{displaymath}
  \forall q\in Q\quad
  f(q) \,\oplus\,
  \bigoplus_{\mathclap{\substack{e\in E\\\mathrm{src}(e)=q}}} w(e) = 1 \, .
\vspace*{-2mm}
\end{displaymath}
An SFST can be brought into locally normalized form (implying global
normalization) using the WFST weight pushing algorithm
\cite[pp.~241--43]{mohri2009weighted}. This is illustrated in
Figure~\ref{fig:push}.

From an SFST $\mathcal{A}$ a random path $\pi$ can be drawn by starting at
state $q_0$, drawing an outgoing edge with probability $w(e)$, and
iterating this process at the next state $\mathrm{tgt}(e)$. At any state
$q$, terminate with probability $f(q)$.

\textbf{Claim:} If $\pi$ is a random path through $\mathcal{A}$, then
$(\mathrm{istr}(\pi), \mathrm{ostr}(\pi))$ is a random string pair
distributed according to $\llbracket\mathcal{A}\rrbracket$.
\emph{Reason:} The morphism $g: E^* \to (\Gamma^*\times\Delta^*)$ defined
for successful paths by $g(\pi) = (\mathrm{istr}(\pi),\mathrm{ostr}(\pi))$
has preimage $g^{-1}(u,v)=\mathrm{Paths}(u,v)$, which is measurable by
(\ref{eq:prob}). By construction, $\pi$ is an arbitrary path in this
countable set of disjoint events and was drawn with probability $w(\pi)$.

In other words, we can sample random string pairs by sampling random
paths. This allows us to gloss over the fact that an SFST $\mathcal{A}$
gives rise to both a probability space over paths and a probability space
over string pairs. We write $X\sim\mathcal{A}$ to refer to a random string
pair with distribution $\llbracket\mathcal{A}\rrbracket$.

It follows that sampling from SFSTs is unaffected by WFST optimizations
which leave the behavior of an automaton intact. This includes weighted
epsilon-removal, determinization, and minimization
\cite{mohri2009weighted}; as well as disambiguation
\cite{mohri2017disambiguation}. This means that $X\sim\mathcal{A}$ and
$Y\sim g(\mathcal{A})$ are equal as random elements, for any combination of
optimizations $g$.  We can in effect sample $Y$ by sampling paths from
$\mathcal{A}$.  As we will see below, this can be efficient in situations
where determinization or disambiguation would lead to an exponential size
increase.

Sampling from SFSTs can be effective even when operations would change
their behavior. Isomorphisms such as FST reversal or inversion are a
particularly simple case: If we want to sample from the reverse of
$\mathcal{A}$, we can either construct and then sample from
$\mathrm{WeightPush}(\mathrm{Reverse}(\mathcal{A}))$; or, equivalently,
sample directly from $\mathcal{A}$ and then reverse the sampled
paths/strings.

More generally, this strategy also works for certain morphisms of string
pairs (such as morphisms of arcs that only change arc labels, including
FST projection) as well as composition with unweighted functional FSTs. For
example, sampling from $\mathrm{Project}_1(\mathcal{A})$ is equivalent to
drawing a random string pair $(u_1,u_2)$ from $\mathcal{A}$ and returning
$(u_1,u_1)$.

\subsection{Epsilon-cycle conflation}

The strategy of sampling random string pairs by sampling random paths can
be inefficient for SFSTs with cyclic epsilon graphs, especially in the
presence of high-probability epsilon cycles. If the probability of the
epsilon-loop at state 1 in Figure~\ref{fig:push} is $1-\delta$, then a
random path will contain $1/\delta - 1$ repetitions of that edge on
average, which can be problematic for small $\delta$, i.e.\ loop
probabilities close to $1$.  Sampling strings by sampling paths is not
efficient in this situation, as the path sampling algorithm has to
repeatedly expand the epsilon-loop despite the fact that it contributes no
symbols to the string labeling.

\begin{figure}
  \centering%
  \includegraphics[clip,trim={36 36 36 36},height=2.5cm]{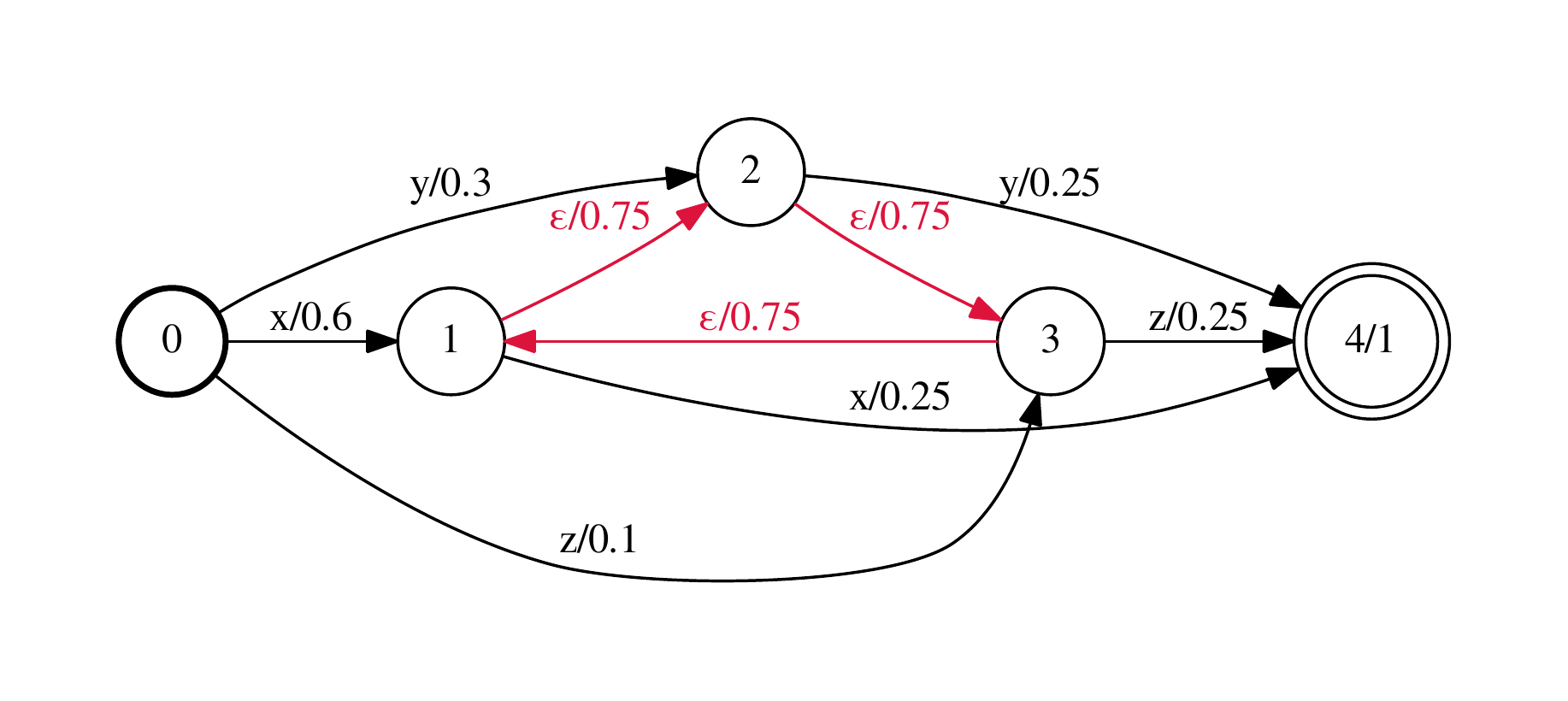}%
  \vspace{-2ex}%
  \caption{Normalized SFST with epsilon-cycle highlighted.}
  \label{fig:epscycle3}%
  \vspace{-4mm}%
\end{figure}

\begin{figure}
  \centering%
  \includegraphics[clip,trim={36 12 36 36},height=3.6cm]{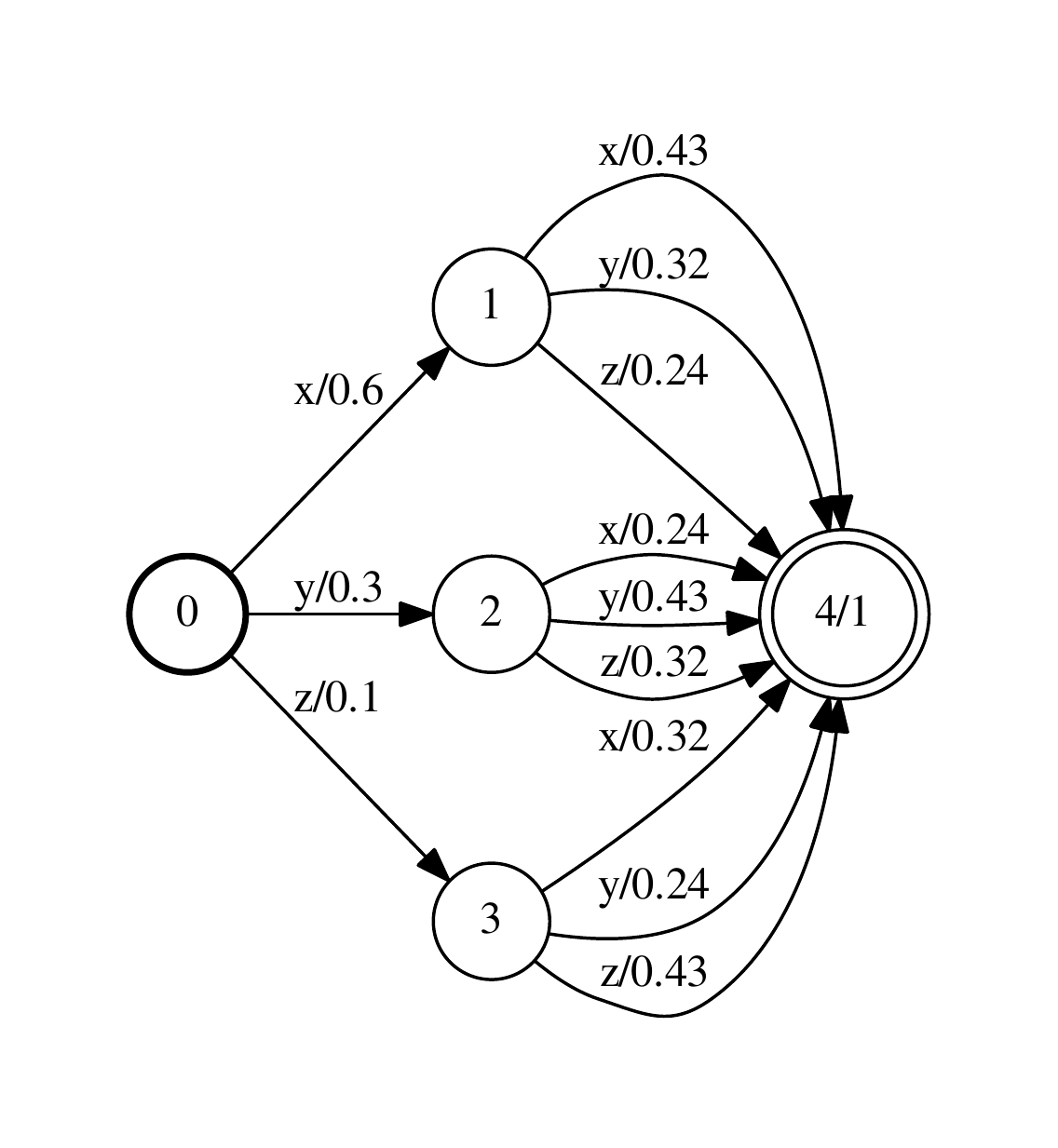}%
  \hspace*{\fill}%
  \includegraphics[clip,trim={36 36 36 36},height=4.1cm]{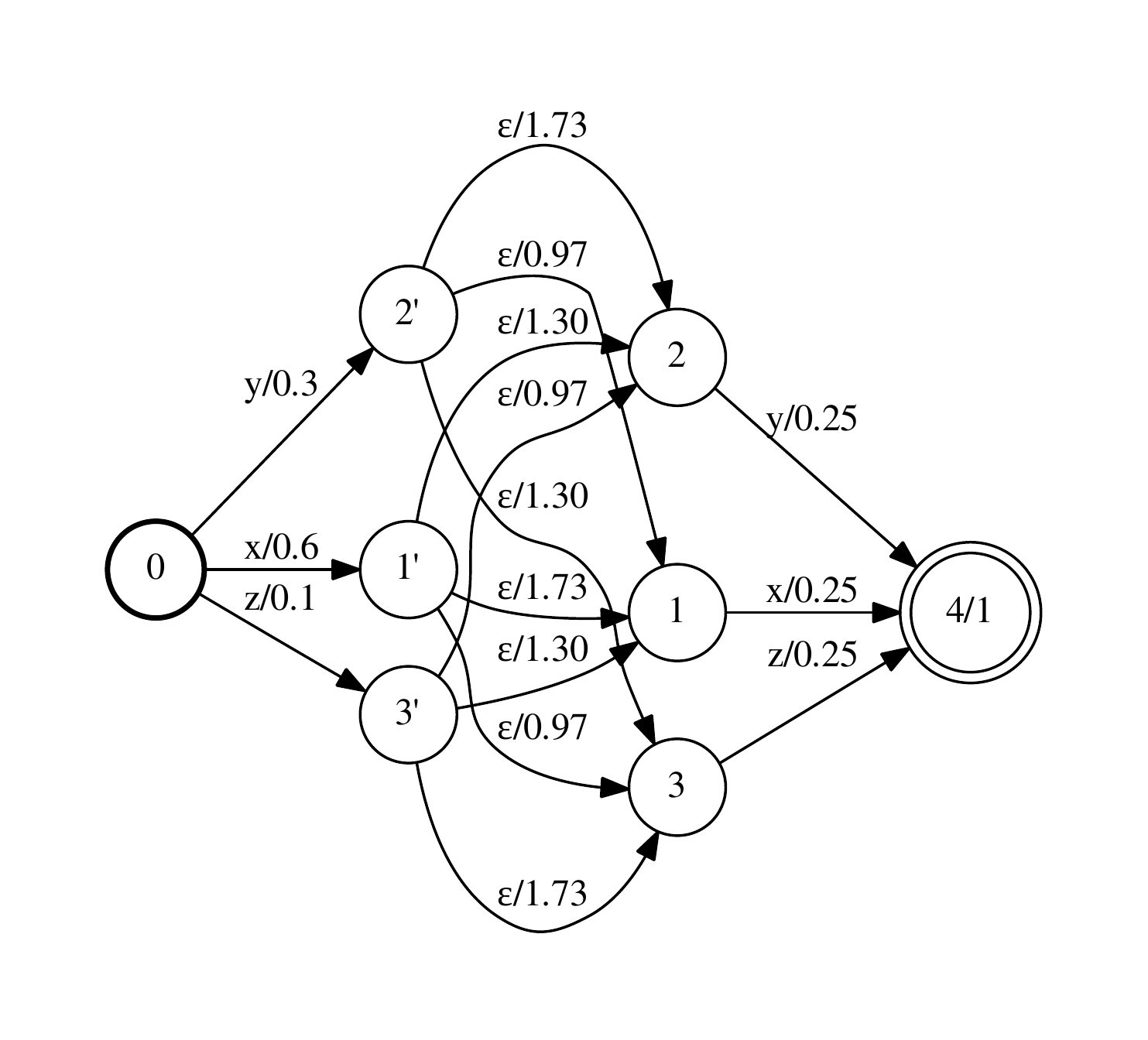}%
  \vspace{-2ex}%
  \caption{SFSTs equivalent to Figure~\ref{fig:epscycle3} after
    epsilon-removal (left) vs.\ epsilon-cycle conflation (right).}
  \label{fig:epscycle4}%
  \vspace{-1mm}%
\end{figure}

One remedy is to apply the weighted epsilon-removal algorithm
\cite[pp.~233--37]{mohri2009weighted}. We illustrate this using an SFST with
a nontrivial strongly-connected component (SCC) of its epsilon-graph,
highlighted in Figure~\ref{fig:epscycle3}. The result of epsilon-removal is
depicted in the left-hand side of Figure~\ref{fig:epscycle4}. Because
epsilon-removal eliminated all epsilon-transitions, new outgoing
transitions were added with appropriate weights for all outgoing
transitions from any state in the original epsilon-SCC.

As we had seen above, in the context of sampling the presence of
epsilon-transitions is not necessarily problematic. It would suffice to
simply ensure that the epsilon-graph is acyclic. This can be achieved by a
modification of the epsilon-removal algorithm, which we call epsilon-cycle
conflation. Rather than removing all epsilon-transitions, this algorithm
merely replaces every set of epsilon-paths through an epsilon-SCC with a
single epsilon-transition. This is illustrated in the right-hand side of
Figure~\ref{fig:epscycle4}; pseudocode appears in
Figure~\ref{alg:rmepscycle}.

The main property of the epsilon-cycle conflation algorithm is that it
splits each state within an epsilon-SCC into two states, by adding a new
state that is paired with an existing state. Transitions from within the
SCC are removed (line~16); transitions from outside the SCC are redirected
to reach the new state (line~18). Each SCC is replaced with a complete
bipartite graph: for each newly split state, epsilon-transitions are added
to all other original states within the SCC (for-loop starting on
line~4). In other words, the bipartite graph that replaces the SCC encodes
the all-pairs algebraic distance within the SCC. The algorithm references
standard subroutines for computing the SCCs (line~2), for computing the
single-source algebraic distance (line~6), and for trimming useless states
(line~21), as well as standard WFST accessor functions for adding and
deleting states and arcs. It also uses a simple arc-filter called
\textsc{EpsilonSccArcFilter} (line 5 and 13), which is a predicate that
returns true iff a given arc is an epsilon-transition within the specified
SCC. It is assumed that the algebraic distance subroutine
\textsc{ShortestDistance} can take this filter as an argument, so as to
restrict it to the specified SCC within the epsilon-graph.\footnote{This
  functionality is readily available in the OpenFst
  library~\cite{allauzen2007openfst}, which forms the basis of our
  implementation.}

\begin{figure}
  \begin{algorithmic}[1]
    \footnotesize%
    \Procedure{ConflateEpsilonCycles}{$\mathcal{A}$}
    \State $\mathit{scc} \gets \textsc{Scc}(\mathcal{A})$
    \Comment{Map from state to SCC number}
    \State $\mathit{split} \gets \textsc{Map}()$
    \Comment{Empty map}
    \For{$(\mathit{state}, \mathit{component})\ \mathbf{in}\ \mathit{scc}$}
    \State $\mathit{filter} \gets \textsc{EpsilonSccArcFilter}(
    \mathit{scc},\ \mathit{component})$
    \State $\mathit{distance} \gets \textsc{ShortestDistance}(
    \mathit{state},\ \mathit{filter})$
    \State $s \gets \textsc{AddState}(\mathcal{A})$
    \For{$(t, c)\ \mathbf{in}\ \mathit{scc}$}
    \If{$c = \mathit{component}$}
    \State $\textsc{AddArc}(\mathcal{A},\
    (s,\ \varepsilon,\ \varepsilon,\ \mathit{distance}[t],\ t))$
    \State $\mathit{split}[\mathit{state}] \gets s$
    \EndIf
    \EndFor
    \EndFor

    \For{$(\mathit{state}, \mathit{component})\ \mathbf{in}\ \mathit{scc}$}
    \State $\mathit{filter} \gets \textsc{EpsilonSccArcFilter}(
    \mathit{scc},\ \mathit{component})$
    \For{$\mathit{arc}\ \mathbf{in}\ \textsc{Arcs}(
      \mathcal{A},\ \mathit{state})$}
    \If{$\mathit{filter}(\mathit{arc})$}
    \State $\textsc{DeleteArc}(\mathcal{A},\ \mathit{arc})$
    \ElsIf{$\mathrm{tgt}(\mathit{arc})\ \mathbf{in}\ \mathit{split}$}
    \State $\mathrm{tgt}(\mathit{arc}) \gets
    \mathit{split}[\mathrm{tgt}(\mathit{arc})]$
    \EndIf
    \EndFor
    \EndFor
    \If{$\textsc{Start}(\mathcal{A})\ \mathbf{in}\ \textit{split}$}
    \State $\textsc{Start}(\mathcal{A}) \gets
    \textit{split}[\textsc{Start}(\mathcal{A})]$
    \EndIf
    \State $\textsc{Connect}(\mathcal{A})$
    \EndProcedure
    \normalsize%
  \end{algorithmic}
  \vspace{-2mm}%
  \caption{Epsilon-cycle conflation algorithm.}%
  \label{alg:rmepscycle}
  \vspace{-4mm}%
\end{figure}

Unlike epsilon-removal, epsilon-cycle conflation only affects the states
and arcs within each epsilon-SCC\@. It does not add or remove any
non-epsilon transitions or epsilon-transitions not fully within
epsilon-SCCs.  If the sizes of the epsilon-SCCs are $S_1, \dots, S_n$,
epsilon-cycle conflation adds at most $\sum_{i=1}^n S_i$ additional states
and no more than $\sum_{i=1}^n S_i^2$ additional transitions. A worst case
arises if the entire FST is the cycle graph $C_m$ with $m$ states and $m$
epsilon-arcs, in which case epsilon-cycle conflation will result in $2m$
states and $m^2$ arcs before trimming.

From here on we assume that any SFST we sample from is locally normalized
and epsilon-cycle free. If it is not, we first apply epsilon-cycle
conflation, followed by weight pushing.

\section{CTC decoding by sampling}

We now turn to an illustration of sampling from stochastic automata in the
context of the decoding problem for Connectionist Temporal Classification
(CTC) \cite{graves2006}. The CTC decoding problem arises because the
sequences generated as CTC outputs do not correspond directly to the final
label sequences. CTC output is a $T\times L$ dimensional matrix $(w_{ij})$
where each vector $w_{i\boldsymbol{\cdot}}$ represents the probability
distribution at time $i$ over symbols from a finite alphabet of labels with
cardinality $L-1$ augmented with a special blank symbol $\varnothing$.

\begin{figure}
\centering
\includegraphics[width=\columnwidth]{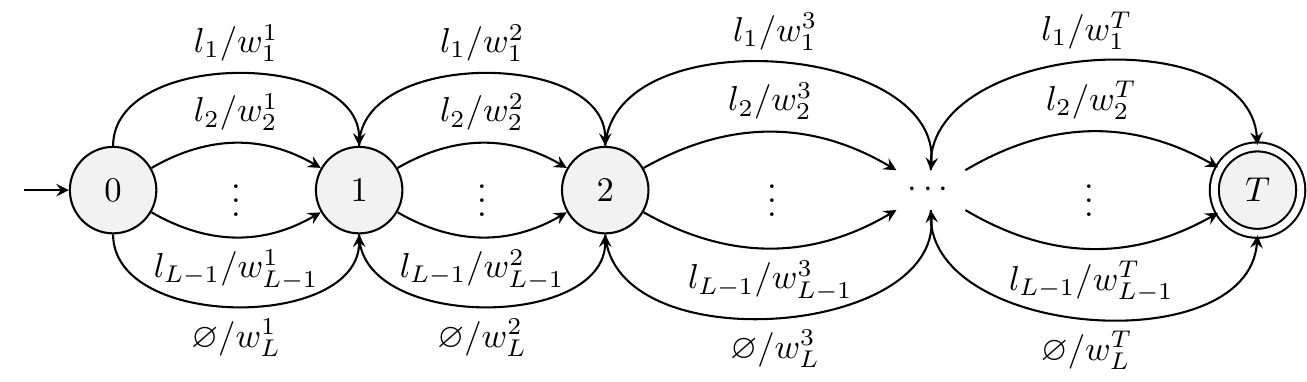}
\caption{CTC predictive distribution as a lattice WFSA $\mathcal{A}$.}
\label{fig:sausage}
\vspace{-5mm}%
\end{figure}

A standard way (e.g.~\cite{miao2015eesen}) of conceptualizing CTC
output is as an equivalent weighted finite state acceptor
(WFSA). Figure~\ref{fig:sausage} shows a WFSA view of a CTC lattice
corresponding to a sequence of length $T$ of symbols from $\{l_1,
\ldots, l_{L-1}, \varnothing\}$.  The cost of transitioning into state $i$
having accepted symbol $l_j$ is $w^i_{\!j}$.

From a path $\pi$ through the CTC lattice we get to a final labeling
sequence $\ell$ by first collapsing contiguous runs of repeated symbols and
then removing blank symbols.  For example, the path labeled with
\textquotedbl$\mathrm{aab}\varnothing\mathrm{b}$\textquotedbl\ gives rise to
the labeling \textquotedbl$\mathrm{abb}$\textquotedbl.  Following
\cite{graves2006} we denote the path-to-labeling function as $\mathcal{B}$.

\begin{figure}
  \centering%
  \includegraphics[clip,trim={36 36 36 36},height=3.6cm]{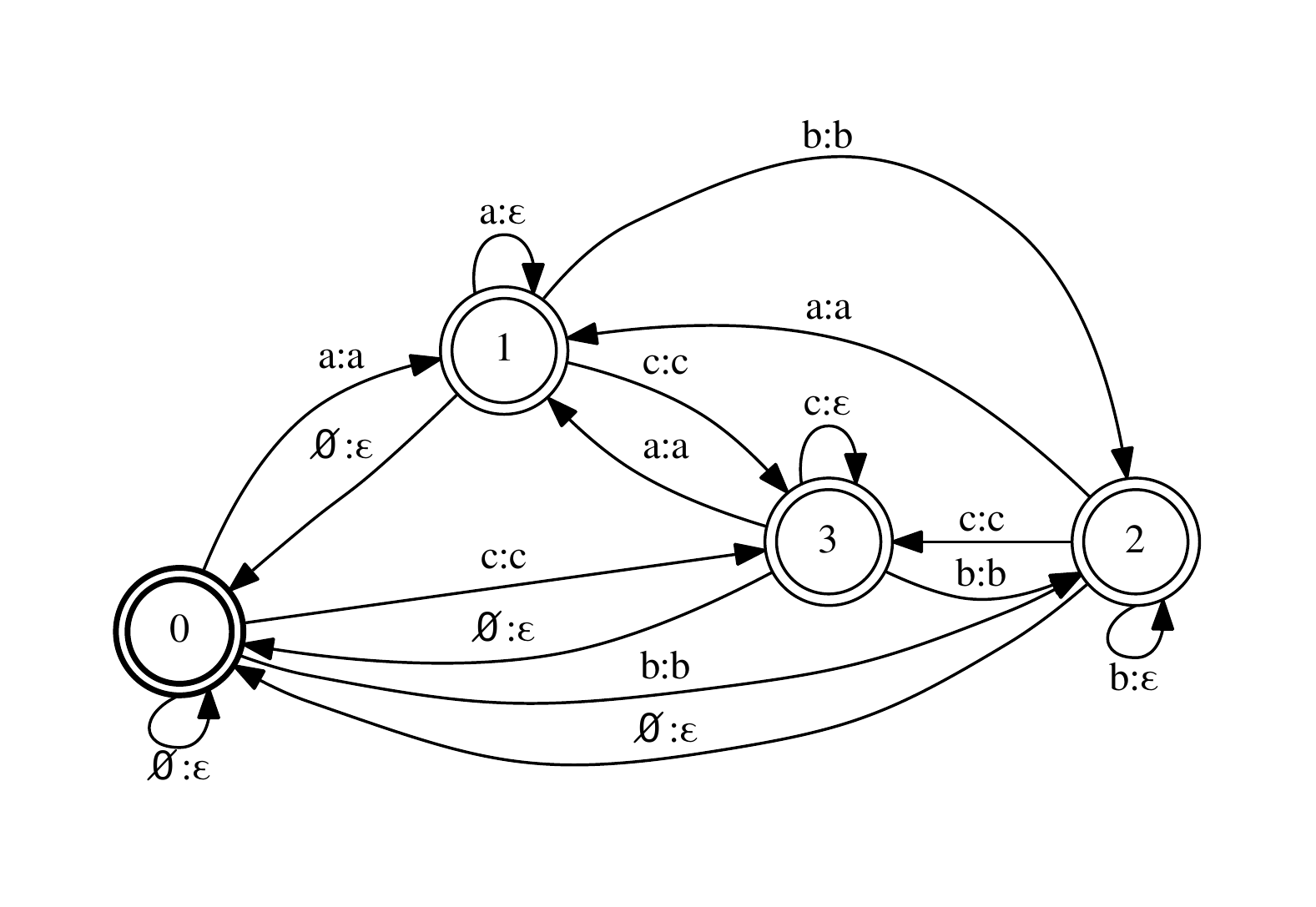}%
  \\%
  \includegraphics[clip,trim={36 36 36 36},width=\columnwidth]{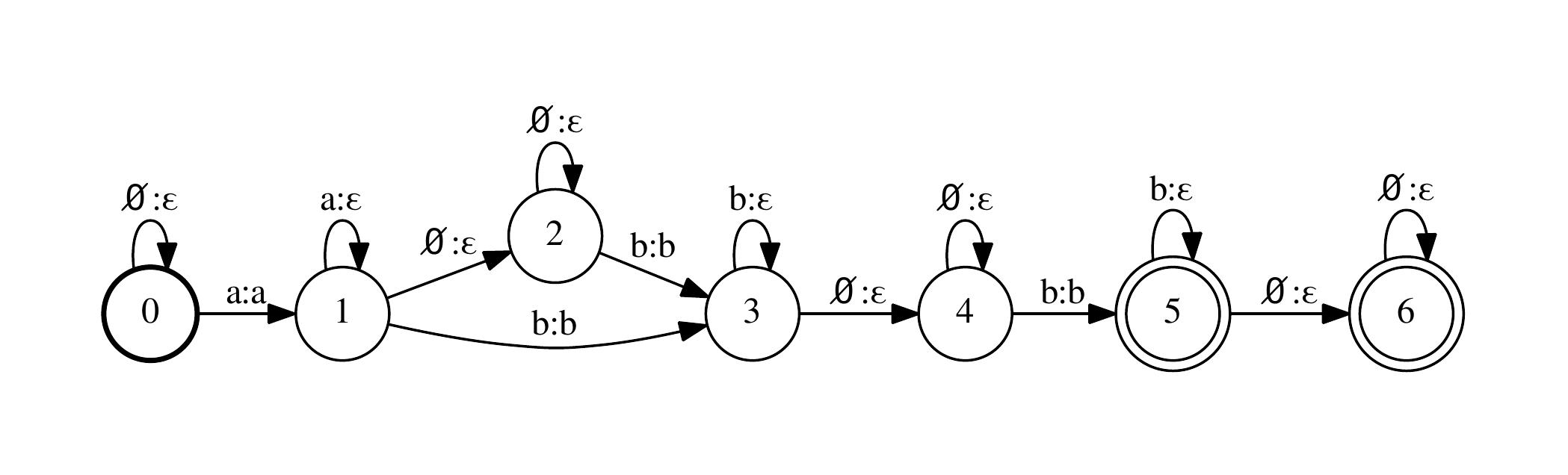}%
  \vspace{-1ex}%
  \caption{Unweighted path-to-labeling FST $\mathcal{B}$ (top) for the
    symbol alphabet $\{\text{\rm a, b, c}\}$ plus blank symbol
    $\varnothing$; and composed FST $\mathcal{B} \circ
    \text{\rm\textquotedbl abb\textquotedbl}$ (bottom) representing
    $\mathcal{B}^{-1}(\text{\rm\textquotedbl abb\textquotedbl})$.}
  \label{fig:path2labeling}%
  \vspace{-5mm}%
\end{figure}

Crucially, the path-to-labeling function $\mathcal{B}$ can be computed by
an unweighted functional FST, which we also refer to as $\mathcal{B}$.  In
order to collapse contiguous runs, the FST only needs to remember the last
symbol seen, i.e.\ it has the structure of a bigram model over its input
labels. This is illustrated in the top FST in
Figure~\ref{fig:path2labeling}, where e.g.\ state~1 is reached whenever the
input label `a' is read in any state. On reaching state~1 from another
state, `a' is also output, but subsequent occurrences of `a' are mapped to
the empty string. A similar construction was used for the \emph{Token WFST}
in~\cite{miao2015eesen}.

For a given SFST $\mathcal{A}$ over a label alphabet $\Gamma$ and
path-to-labeling transducer $\mathcal{B}$, the posterior predictive
distribution over labeling sequences $\ell\in \Gamma^*$ corresponds to the
output-projection of the WFST composition $\mathcal{A}\circ\mathcal{B}$:
\begin{equation}
  p(\ell) =
  \llbracket\mathrm{Project}_2(\mathcal{A} \circ \mathcal{B})\rrbracket(\ell, \ell)
  \, .
  \label{eq:predictive}
\end{equation}
In maximum a~posteriori (MAP) decoding we are asked to find
$
  \ell^\star = \mathrm{argmax}_{\ell\in \Gamma^*}\ p(\ell)
$.
It is easy to define more complex decoding tasks over the labeling
distribution $p$ from~(\ref{eq:predictive}).

The key point of this paper is that we can sample efficiently from the
labeling distribution $p$ for many choices of $\mathcal{A}\!$ and $\mathcal{B}$,
including but not limited to CTC decoding, where exact MAP decoding via
disambiguation or determinization is generally only tractable for
unrealistically small WFSTs. When $(\ref{eq:predictive})$ involves only
operations discussed in Section~\ref{sec:intro} above, we know that
path-sampling is effective. All we have to do is sample paths from the SFST
$\mathcal{A}\!\!,$ pass them through the unweighted functional FST
$\mathcal{B}$, and read labeling strings off the output tape.

This immediately leads us to a naive decoding strategy: Sample a large
number of labeling strings and return the labeling which occurs most
frequently in the sample, i.e.\ the sample mode. The advantage is that this
is very easy to implement. On the other hand, the naive strategy often
needs to sample a large number of paths, both to ensure that there are no
unexplored modes and to distinguish labelings that are nearly equally
likely.

The naive strategy can be improved by evaluating the probability $p(\ell)$
from~(\ref{eq:predictive}) for a given labeling $\ell$ or set of labelings
$Y$. We do this by expressing the preimage $\mathcal{B}^{-1}(Y) = \{x \mid
\mathcal{B}(x)\in Y\}$ as a WFST $\mathcal{B}\circ Y$ (see the bottom FST
in Figure~\ref{fig:path2labeling}) and measuring that preimage using
$\mathcal{A}$. For singleton $Y=\{\ell\}$ the number of states and arcs of
the preimage FST is linear in $|\ell|$ by construction. Evaluating
$p(\ell)$ amounts to computing the grand total weight from~(\ref{eq:prob})
of $\mathcal{A}\circ(\mathcal{B}\circ\ell)$ using standard WFST
operations. It is important to note that this efficiently sums over an
implicitly represented WFST that compactly encodes an exponentially larger
set of paths. This means that, while we could compute lower bounds for
$p(\ell)$ via sampling and path probabilities alone, those bounds are so
loose as to be of little practical value, due to the large number of paths
for the same labeling.

The ability to evaluate (\ref{eq:predictive}) is useful in several
ways. For example, if we know that $p(\ell)>0.5$ for a particular labeling
$\ell$, then $\ell$ must necessarily be the mode of $p$. Generalizing from
here gives us our sampling-based CTC decoding strategy (pseudocode appears
in Figure~\ref{alg:ctcdecode1}): Repeatedly draw $\ell$ using path-based
sampling (lines 12--13), compute $p(\ell)$ (line 19) and track the most
probable labeling seen (lines 3--4 and 21--22) as well as the probability
mass of all seen labelings (lines 10 and 20). We always include the
labeling for the most likely path (line 2), a fast and useful heuristic already noted in
\cite{graves2006}.  If the probability of the most probable seen labeling
exceeds the unseen probability mass (lines 5 and 23), then that labeling
must be the global mode.  For efficiency the decoding procedure takes parameters
specifying a maximum number of random draws and a threshold $\theta$ for
early stopping.

There are many ways to conceptualize when to stop sampling (line 25), most
generally as an optimal stopping problem \cite{ferguson2008optimal}, which
subsumes sequential probability ratio tests. We want to stop sampling when
we are reasonably certain that we have already seen the global mode.%
\footnote{%
  The early stopping strategies are applicable because we can draw iid
  samples from the posterior predictive distribution. If instead we sampled
  correlated variates sequentially, there would be a substantial risk that
  e.g.\ a Markov chain sampler might get stuck near a local maximum.}
Since in this strategy we can observe the probabilities of labelings, we could
hypothesize that there is an unseen mode that occurs with probability
$P_0$, which is a random variable that we assume follows a
$\mathrm{Beta}(1, n+1)$ distribution. The probability $P_0$ of the unseen
mode would have to exceed the highest labeling probability $p^\star$
encountered so far, and cannot exceed the unseen probability mass $1-t$. We
can stop sampling as soon as this becomes sufficiently unlikely, when
$\Pr(p^\star \leq P_0 \leq 1-t) = (1 - p^\star)^{n+1} - t^{n+1} < \theta$.

\begin{figure}
  \begin{algorithmic}[1]
    \footnotesize%
    \Function{CtcDecode}{$\mathcal{A},\ \mathcal{B},\ \mathit{max\_draws},\ \theta$}
    \State $\pi^\star \gets \textsc{ShortestPath}(\mathcal{A})$ \Comment{most likely path}
    \State $\ell^\star \gets \textsc{RmEpsilon}(\textsc{Project}_2(\pi^\star\circ\mathcal{B}))$
    \State $p^\star \gets \textsc{ShortestDistance}(\mathcal{A} \circ (\mathcal{B} \circ \ell^\star))$
    \If{$p^\star > 0.5$}
    \State \textbf{return} $\ell^\star$ \Comment{global mode found}
    \EndIf
    \State $C \gets \textsc{Map}()$ \Comment{empty map of labelings to counts}
    \State $C[\ell^\star] \gets 1$
    \State $S \gets \{\ell^\star\}$ \Comment{set of labelings with known probability}
    \State $t \gets p^\star$ \Comment{total seen probability mass}
    \For{$n \gets 1 \ldots \mathit{max\_draws}$}
    \State $\pi \gets \textsc{RandGen}(\mathcal{A})$ \Comment{random path}
    \State $\ell \gets \textsc{RmEpsilon}(\textsc{Project}_2(\pi \circ \mathcal{B}))$
    \If{$\ell \notin C$}
    \State $C[\ell] \gets 0$
    \EndIf
    \State $c \gets C[\ell] \gets C[\ell] + 1$
    \If{$\ell\notin S$ \textbf{and} $\textsc{ComputeProbability}(c,\ n,\ p^\star\!\!\!,\ t,\ \theta)$}
    \State $S \gets S \cup \{\ell\}$
    \State $p \gets \textsc{ShortestDistance}(\mathcal{A} \circ (\mathcal{B} \circ \ell))$
    \State $t \gets t + p$
    \If{$p > p^\star$}
    \State $\ell^\star, p^\star \gets \ell, p$
    \EndIf
    \If{$p^\star > 1 - t$}
    \State \textbf{break} \Comment{global mode found}
    \EndIf
    \EndIf
    \If{$\Pr(p^\star \leq P_0 \leq 1-t) < \theta$}
    \Comment{for $P_0 \sim \mathrm{Beta}(1, n+1)$}
    \State \textbf{break} \Comment{approximate early stopping}
    \EndIf
    \EndFor
    \State \textbf{return} $\ell^\star$
    \EndFunction
    \normalsize%
  \end{algorithmic}
  \vspace*{-2mm}%
  \caption{Sampling-based CTC decoding algorithm.}%
  \label{alg:ctcdecode1}
  \vspace{-5mm}%
\end{figure}

While the evaluation of labeling probabilities provides useful information
that guides early stopping, it is often not necessary to compute $p(\ell)$
for every labeling string sampled randomly. Like the naive sampling
strategy, our decoding algorithm keeps a count of labelings (lines 7--8 and
14--16). The count $c$ (line 16) of the current labeling $\ell$, together
with other information, is passed to a predicate
$\textsc{ComputeProbability}$ indicating whether $p(\ell)$ should be
evaluated during a given iteration (the \emph{probability computation strategy}),
which can be implemented in a variety of ways. If it is
always true, probabilities are computed for every distinct labeling, which
can improve early stopping. If it is always false and if the most frequent
labeling in $C$ is returned, this becomes the naive sampling algorithm.
$\textsc{ComputeProbability}$ could also be implemented (similar to the negation of the early stopping criterion) as $\Pr(p^\star \leq P_c \leq 1-t) \geq \theta$
where $P_c$ is a random variable with a $\mathrm{Beta}(c+1, n-c+2)$
distribution, so we compute $p(\ell)$ only when this value is likely to
exceed $p^\star$. Finally, the simple criterion $c>1$ works well in our
experience: only evaluate the probability of a labeling when seeing it
for the second time.

The decoding algorithm can thus be fine-tuned for many practical
considerations, and many additional variations are possible, such as
sampling in batches and only occasionally testing for early stopping. In
the following section we compare different decoding strategies, including
the decoding algorithm from Figure~\ref{alg:ctcdecode1} and the naive
sampling algorithm.


\section{Evaluation and discussion}

Our experiments focus on continuous phoneme recognition, which is a
sequence-to-sequence task where the input sequence corresponds to the
acoustic features and the output is a sequence of categorical labels
representing phonemes from a finite inventory. CTC-based approaches have
been successfully applied to this
task~\cite{fernandez2008,graves2013,sak2015,pundak2016}. We trained a CTC
phoneme recognizer for Argentinian Spanish\footnote{Our corpus is
  available at \url{http://openslr.org/61/}.}
based on the architecture
described in~\cite{adams2018}. We computed CTC output matrices
on a test set of 90 unseen utterances and turned those
into CTC lattices.

\begin{table}
  \centering%
  \caption{Comparison of decoding strategies in terms of success at finding
    the global mode and mean number of paths sampled.}%
  \label{tab:eval}%
  \vspace{-2mm}%
  \scriptsize%
  \rowcolors{3}{light-gray}{white}%
  \setlength\tabcolsep{2pt}
  \def\arraystretch{1.2}
  \begin{tabular}{rlrrr}
    \toprule
    & Decoding strategy, & Mode & Avg.~paths & Avg.~probs. \\
    & probability computation strategy & found & sampled & computed \\
    \midrule
    1 & $\textsc{BeamSearch}(100)$, n/a                   &  83\% &      &    \\
    2 & $\textsc{BeamSearch}(2000)$, n/a                  &  83\% &      &    \\
    3 & $\textsc{NaiveSampling}(600,\ \theta=0)$, never   &  82\% &  600 &  0 \\
    4 & $\textsc{NaiveSampling}(6000,\ \theta=0)$, never  &  94\% & 6000 &  0 \\
    5 & $\textsc{CtcDecode}(0,\ \theta=0)$, never         &  77\% &    0 &  0 \\
    6 & $\textsc{CtcDecode}(100,\ \theta=0.01)$, always   &  99\% &   36 & 27 \\
    7 & $\textsc{CtcDecode}(600,\ \theta=0.01)$, always   & 100\% &   53 & 40 \\
    8 & $\textsc{CtcDecode}(600,\ \theta=0.01)$, if $c>1$ & 100\% &   53 &  7 \\
    \bottomrule
  \end{tabular}
  \normalsize%
  \vspace{-4mm}%
\end{table}

We compared different decoding strategies in terms of their ability to find
the global mode of the posterior predictive labeling distribution. We first
ran an exhaustive search to locate the global mode. We then compared
different decoding strategies in terms of how often they were able to
locate the global mode. The results are summarized in Table~\ref{tab:eval}.
The first two rows refer to the CTC beam search decoding algorithm of
TensorFlow~\cite{abadi2016}, which we ran with its default beam-width of
$100$ and then again with a much larger beam-width of $2000$, which did not
affect the results of beam search.

The third and fourth row of Table~\ref{tab:eval} show results for the naive
sampling strategy, which simply returns the mode of a large sample. Because
no early stopping is employed, the average number of paths sampled per
utterance is identical to the $\mathit{max\_draws}$ parameter of
Figure~\ref{alg:ctcdecode1}. In the naive sampling strategy the number of
probability computations is zero by design. This strategy can match or
exceed beam-search in its ability to locate the global mode, but this comes
at a substantial cost, as a large number of paths needs to be sampled. We
experimented with early stopping for naive sampling before deciding against
it: while early stopping can ensure that there are no unseen modes, a
reduction in sample size made it much harder to correctly identify the global
mode within the sample.

The fifth row of Table~\ref{tab:eval} is the best-path heuristic mentioned
in \cite{graves2006}. It only considers the labeling of the most likely
path $\pi^\star$ and does not sample any additional paths. While it
performs worst in this comparison, it is a useful heuristic due to speed.

The last three rows of Table~\ref{tab:eval} are nontrivial instances of our
decoding algorithm. We fixed a confidence threshold of
$\theta=0.01$. Larger values result in fewer overall computations, at the
expense of possibly not finding the global mode. Smaller values increase
the overall computations. We then chose a maximum number of $600$ draws,
which always leads to early stopping: the algorithm from
Figure~\ref{alg:ctcdecode1} either finds the global mode (lines 6 and 24)
or stops early with confidence $\theta$ (line 26). When running with a
sample limit of $600$ paths, our algorithm always found the global mode in
our experiments. Even with a much smaller limit of at most $100$ random
draws, our algorithm still did well. Finally, as the last row shows, the
average number of probability evaluations can be greatly reduced by
computing labeling probabilities only when a labeling is encountered more
than once. The last three strategies shown here all outperform beam search.

\section{Conclusion}

We have discussed sufficient conditions under which sampling random paths
from transformations of stochastic automata yields random strings from the
transformed automata. We have shown that this can be used to sample
labeling sequences from CTC posterior lattices. In a comparison on a
connected phoneme recognition problem, sampling-based decoding informed by
posterior probabilities was able to locate posterior modes reliably with
minimal search effort.

\bibliographystyle{IEEEtran}
\bibliography{ctc_sampling,fst}

\begin{thebibliography}{10}
\providecommand{\url}[1]{#1}
\csname url@samestyle\endcsname
\providecommand{\newblock}{\relax}
\providecommand{\bibinfo}[2]{#2}
\providecommand{\BIBentrySTDinterwordspacing}{\spaceskip=0pt\relax}
\providecommand{\BIBentryALTinterwordstretchfactor}{4}
\providecommand{\BIBentryALTinterwordspacing}{\spaceskip=\fontdimen2\font plus
\BIBentryALTinterwordstretchfactor\fontdimen3\font minus
  \fontdimen4\font\relax}
\providecommand{\BIBforeignlanguage}[2]{{%
\expandafter\ifx\csname l@#1\endcsname\relax
\typeout{** WARNING: IEEEtran.bst: No hyphenation pattern has been}%
\typeout{** loaded for the language `#1'. Using the pattern for}%
\typeout{** the default language instead.}%
\else
\language=\csname l@#1\endcsname
\fi
#2}}
\providecommand{\BIBdecl}{\relax}
\BIBdecl

\bibitem{mohri2009weighted}
M.~Mohri, ``Weighted automata algorithms,'' in \emph{Handbook of Weighted
  Automata}, M.~Droste, W.~Kuich, and H.~Vogler, Eds.\hskip 1em plus 0.5em
  minus 0.4em\relax Springer, 2009, pp. 213--254.

\bibitem{mohri2017disambiguation}
M.~Mohri and M.~D. Riley, ``A {D}isambiguation {A}lgorithm for {W}eighted
  {A}utomata,'' \emph{Theoretical Computer Science}, vol. 679, pp. 53--68,
  2017.

\bibitem{allauzen2007openfst}
C.~Allauzen, M.~Riley, J.~Schalkwyk, W.~Skut, and M.~Mohri, ``{O}pen{F}st: {A}
  {G}eneral and {E}fficient {W}eighted {F}inite-{S}tate {T}ransducer
  {L}ibrary,'' in \emph{CIAA 2007: Implementation and Application of Automata},
  2007, pp. 11--23.

\bibitem{graves2006}
A.~Graves, S.~Fern{\'a}ndez, F.~Gomez, and J.~Schmidhuber, ``Connectionist
  {T}emporal {C}lassification: {L}abelling {U}nsegmented {S}equence {D}ata with
  {R}ecurrent {N}eural {N}etworks,'' in \emph{Proc. 23rd International
  Conference on Machine Learning (ICML)}, Pittsburgh, Pennsylvania, Jun. 2006,
  pp. 369--376.

\bibitem{miao2015eesen}
Y.~Miao, M.~Gowayyed, and F.~Metze, ``{EESEN}: End-to-end speech recognition
  using deep {RNN} models and {WFST}-based decoding,'' in \emph{Automatic
  Speech Recognition and Understanding Workshop (ASRU)}, Scottsdale, Arizona,
  Dec. 2015.

\bibitem{ferguson2008optimal}
T.~S. Ferguson, ``Optimal stopping and applications,'' 2008,
  \url{http://www.math.ucla.edu/~tom/Stopping/}.

\bibitem{fernandez2008}
S.~Fern{\'a}ndez, A.~Graves, and J.~Schmidhuber, ``Phoneme recognition in
  {TIMIT} with {BLSTM}-{CTC},'' \emph{arXiv preprint arXiv:0804.3269}, 2008.

\bibitem{graves2013}
A.~Graves, A.~R. Mohamed, and G.~Hinton, ``Speech recognition with deep
  recurrent neural networks,'' in \emph{2013 IEEE International Conference on
  Acoustics, Speech and Signal Processing (ICASSP)}, Canada, May 2013, pp.
  6645--6649.

\bibitem{sak2015}
H.~Sak, F.~de~Chaumont~Quitry, T.~Sainath, K.~Rao \emph{et~al.}, ``Acoustic
  modelling with {CD-CTC-SMBR} {LSTM} {RNN}s,'' in \emph{2015 IEEE Workshop on
  Automatic Speech Recognition and Understanding (ASRU)}, Scottsdale, Arizona,
  Dec. 2015, pp. 604--609.

\bibitem{pundak2016}
G.~Pundak and T.~N. Sainath, ``Lower {F}rame {R}ate {N}eural {N}etwork
  {A}coustic {M}odels,'' in \emph{Interspeech 2016}, San Francisco, USA, Sep.
  2016, pp. 22--26.

\bibitem{adams2018}
O.~Adams, T.~Cohn, G.~Neubig, H.~Cruz, S.~Bird, and A.~Michaud, ``Evaluating
  phonemic transcription of low-resource tonal languages for language
  documentation,'' in \emph{Proceedings of the Eleventh International
  Conference on Language Resources and Evaluation (LREC 2018)}, Japan, 2018,
  pp. 3356--3365.

\bibitem{abadi2016}
M.~Abadi, P.~Barham, J.~Chen, Z.~Chen, A.~Davis, J.~Dean, M.~Devin,
  S.~Ghemawat, G.~Irving, M.~Isard \emph{et~al.}, ``{TensorFlow}: A {S}ystem
  for {L}arge-{S}cale {M}achine {L}earning,'' in \emph{Proc. 12th USENIX
  Symposium on Operating Systems Design and Implementation (OSDI '16)}, 2016,
  pp. 265--283.

\end{thebibliography}

\end{document}